\documentclass[conference]{IEEEtran}
\IEEEoverridecommandlockouts
\usepackage{cite}
\usepackage{amsmath,amssymb,amsfonts}
\usepackage{algorithmic}
\usepackage{graphicx}
\usepackage{textcomp}
\usepackage{xcolor}
\usepackage{times}
\usepackage{slashbox}
\usepackage{latexsym}
\usepackage{times,epsfig}
\usepackage{url}
\usepackage{stmaryrd}
\usepackage{multicol,lipsum}
\usepackage{epsfig,subfigure,bm,dsfont}
\usepackage{latexsym}
\usepackage{dcolumn}
\usepackage{epsf}
\usepackage{float}
\usepackage{mathptmx}

\usepackage{microtype}

\usepackage{fancyhdr}

\def\BibTeX{{\rm B\kern-.05em{\sc i\kern-.025em b}\kern-.08em
    T\kern-.1667em\lower.7ex\hbox{E}\kern-.125emX}}
\begin{document}

\title{Dense Embeddings Preserving the Semantic Relationships in WordNet\\
\thanks{© 2022 IEEE. This paper is accepted at IEEE International Joint Conference on Neural Networks (IJCNN) 2022. This is the preprint version.}
}

\author{\IEEEauthorblockN{1\textsuperscript{st} Canlin Zhang}
\IEEEauthorblockA{\textit{Circulo Health (present)} \\
\textit{Department of Mathematics} \\
\textit{Florida State University (sponsor)} \\
Columbus, US \\
canlin.zhang@circulohealth.com}
\and
\IEEEauthorblockN{2\textsuperscript{nd} Xiuwen Liu}
\IEEEauthorblockA{\textit{Department of Computer Science} \\
\textit{Florida State University}\\
Tallahassee, US \\
liux@cs.fsu.edu}}

\maketitle
\thispagestyle{fancy}

\begin{abstract}
  In this paper, we provide a novel way to generate low dimensional vector embeddings for the noun and verb synsets in WordNet, where the hypernym-hyponym relationship is preserved in the embeddings. We call this embedding the Sense Spectrum (and Sense Spectra for embeddings). In order to create suitable labels for the training of sense spectra, we designed a new similarity measurement for noun and verb synsets in WordNet. We call this similarity measurement the Hypernym Intersection Similarity (HIS), since it compares the common and unique hypernyms between two synsets. Our experiments show that on the noun and verb pairs of the SimLex-999 dataset, HIS outperforms the three similarity measurements in WordNet. Moreover, to the best of our knowledge, the sense spectra provide the first dense synset embeddings that preserve the semantic relationships in WordNet.
\end{abstract}

\begin{IEEEkeywords}
Knowledge Representation, WordNet, Semantic Relationship, Embeddings 
\end{IEEEkeywords}

\section{Introduction}
  WordNet is a lexical database for the English language \cite{WordNet_introduction}, which groups English words into sets of synonyms called $synsets$ \cite{WordNet2}. Each synset is related to a specific semantic sense, and synsets related to the same semantic sense are usually ordered by their usage frequencies in English. There are four syntactic types of synsets in WordNet: noun (n), verb (v), adjective (a) and adverb (r). As a result, a synset in WordNet is represented in the form of ``semantic sense.syntactic type.ordering". For instance, $domestic\_animal.n.01$ means the first noun synset related to the semantic sense ``domestic animal", and $eat.v.03$ means the third verb synset related to the semantic sense ``eat". 
  
  
WordNet can be regarded as a dictionary, since it provides short definitions and usage examples for each synset. On the other hand, WordNet can also be regarded as a thesaurus \cite{WordNet4}, since it records a number of semantic relationships among synsets or their members (called $lemmas$). The most important relationship among synsets in WordNet is the Hypernym-Hyponym relationship \cite{Yamada_2009_hyper}, which indicates the generic term (hypernym) and a specific instance of it (hyponym). Only noun and verb synsets in WordNet possess the hypernym-hyponym relationship \cite{WordNet3}.

Since almost all the state-of-the-art Natural Language Processing (NLP) models are built on embeddings \cite{mikolov2013distributed, Devlin_BERT}, it is desirable to represent the synsets in WordNet by embeddings as well. 
To be specific, low dimensional embeddings (dense embeddings) that can preserve the semantic relationships among synsets in WordNet are especially desired, which  have not been realized. However, we note that the noun and verb synsets make up more than 80 percent of the synsets in WordNet, and the hypernym-hyponym relationship is the major relationship for nouns and verbs. Hence, it should be a good start of the research if we generate dense embeddings that preserve the hypernym-hyponym relationship for noun and verb synsets in WordNet. In this paper, we will not work with adjective or adverb synsets since they do not have the hypernym-hyponym relationship.

We first create a new similarity measurement called the Hypernym Intersection Similarity (HIS), by which the ``commonness" and ``differences" between two noun or verb synsets are measured according to the intersection situation of their hypernym sets. Then, using HIS as labels, we train our embedding vectors with a novel operation other than the inner product, which makes our embedding vectors look like a ``spectrum of senses." So, we call it the Sense Spectrum. 

In the next section, we shall discuss the related work on creating embeddings for WordNet synsets. Then in Section 3, we shall introduce the architectures of our model. In Section 4, we will describe our implementations and provide experimental results. Then in Section 5, we will provide further discussions on our model. Finally,we will conclude the paper with a brief summary in Section 6.

\section{Related Work}
 Synset embeddings have been applied to improve the performances of  NLP models on NLP tasks. For example, \cite{Recski_SimLex_999} combine word embeddings and synset embeddings to improve the performance on measuring the semantic similarity of words. Also, \cite{lesk_algorithm} apply a Lesk algorithm model \cite{Lesk_initial} to do Word Sense Disambiguation (WSD) based on the embeddings of WordNet synsets. Almost all the NLP models evolved with WordNet represent synsets by embedding vectors.


Roughly speaking, there are two ways to create embeddings for WordNet synsets: One way is to combine the embeddings of words appeared in the definition or usage examples of that synset, where pre-trained word embeddings from other models are required \cite{Autoextend}. Another way is to keep each synset in one unique dimension when creating embeddings, and then create a binary matrix recording the existence (or not) of one specific semantic relationship between any two synsets (two dimensions) \cite{WordNet_Embeddings}. 

Synset embeddings created in the first way are dense and low dimensional, yet preserve no semantic relationships. This is because words in the definition and usage examples of a synset barely contain the information on semantic relationships. Synset embeddings created in the second way do preserve the semantic relationships, but these embeddings are sparse and have extremely high dimensions (10,000 dimensions or higher). When being used as neural network inputs, sparse and high dimensional embeddings require more weights in the low layers of neural networks, which often lead to over-fitting \cite{cursedim}. Hence,  low dimensional embeddings that can preserve the semantic relationships among synsets in WordNet are desired, which motivates the proposed sense spectrum.

\section{Architectures}
In the first subsection, we shall introduce the proposed HIS Similarity. Then, in the next subsection, we shall introduce the three basic similarity measurements in WordNet, which will be used as comparisons to our HIS similarity. After that, the formulas and training algorithms of sense spectra will be given.

Besides, we note that it is not very  meaningful to compare a noun synset with a verb one. So, whenever we mention ``two (noun or verb) synsets $a$ and $b$" in this paper, we assume that either both $a$ and $b$ are noun synsets, or both of them are verb ones. 

\subsection{Hypernym intersection similarity}

Primarily, we note that WordNet not only provides the direct hypernym for each noun and verb synset, but also provides its $\mathbf{hypernym \ closure}$ \cite{WordNet_introduction}: Suppose $h_1$ is a direct hypernym of the synset $a$, and $h_2$ is a direct hypernym of $h_1$. Then, the hypernym closure of $a$ will contain both $h_1$ and $h_2$. That is, the hypernym closure consists of ``all the hypernyms of all the hypernyms" for the synset $a$, which is denoted as $H_a$ in this paper. 

For example, if we set synset $a$ to be $man.n.01$ and synset $b$ to be $woman.n.01$, their hypernym closures $H_a$ and $H_b$ are then shown in Figure 1:
\begin{figure}[H]
\centering
\includegraphics[width=8cm]{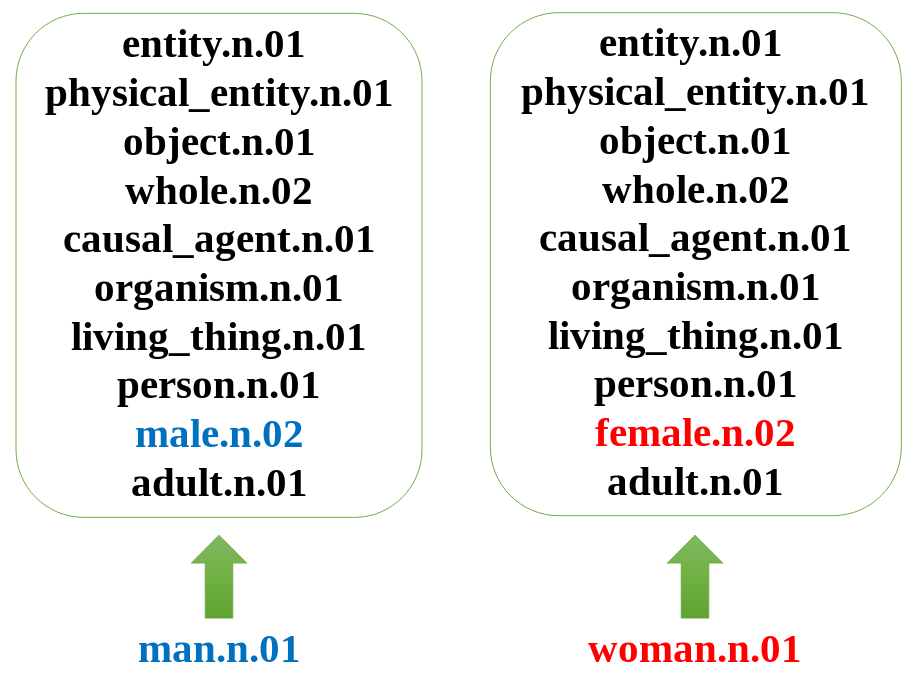}
\caption{(To be viewed in color) The hypernym closures of synsets $man.n.01$ and $woman.n.01$.}
\end{figure} 

The synset $man.n.01$ denotes the common sense of ``man", whose definition in WordNet is ``An adult person who is male (as opposed to a woman)". Accordingly, $woman.n.01$ is defined as ``an adult female person (as opposed to a man)". They have the same direct hypernym $adult.n.01$. We can see from Figure 1 that all the hypernyms of $man.n.01$ and $woman.n.01$ are the same, except that $male.n.02$ is unique to $man.n.01$ and $female.n.02$ is unique to $woman.n.01$. Hence, the hypernym closures are the key to describe the hypernym-hyponym relationship between two synsets $a$ and $b$. 

However, we will not use the hypernym closure directly in the HIS Similarity. This is because the semantic field of a synset should be smaller than that of its hypernym closure \cite{Gao_2013_semantic}. Hence, based on the hypernym-hyponym relationship, the precise representation of a synset should be its hypernym closure plus the synset itself, which is the $\mathbf{hypernym \ set}$ $S_a=H_a\cup \{a\}$. We shall build the HIS Similarity based on the hypernym set.




 Then, for two synsets $a$ and $b$, we define the ``commonness" between them as $S_a\cap S_b$, which can also be denoted as $S_{a\cap b}$. And the ``uniqueness" of synset $a$ is defined as $S_a/S_{a \cap b}$, which consists of the hypernyms unique to the synset $a$ (the ones not in $S_{a \cap b}$). We denote $S_a/S_{a \cap b}$ as $S_{a/b}$. Similarly, the ``uniqueness" of synset $b$ is defined as $S_{b/a}={S_b/S_{a\cap b}}$. We call $S_{a\cap b}$, $S_{a/b}$ and $S_{b/a}$ the Hypernym Representation Sets. 

Taking $a=man.n.01$ and $b=woman.n.01$ as our example again, we can see from Figure 1 that
\begin{align*}
S_{a\cap b}&\!=\!\{adult\!.n.01\!, \ person.n.01\!, \ living\!\_thing.n.01\!,\\
 &organism.n.01\!, \ causal\!\_agent.n.01\!, \ whole.n.02\!,\\
 &object.n.01\!, \ physical\!\_entity.n.01\!, \ entity.n.01\!\},\\
S_{a/b}&=\{man.n.01, \ male.n.02\},\\ 
S_{b/a}&=\{woman.n.01, \ female.n.02\}
\end{align*}

Then, we set $\alpha=|S_{a/b}|$, $\beta=|S_{b/a}|$ and $\gamma =|S_{a\cap b}|$, which denotes the size of each hypernym representation set. Again, if $a=man.n.01$ and $b=woman.n.01$, we have that $\alpha=\beta=2$ and $\gamma=9$.

Finally, for any two noun or verb synsets $a$ and $b$, the $\mathbf{Hypernym \ Intersection}$ $\mathbf{Similarity}$ (HIS) is defined as:
\begin{align}
\mathcal{K}_{a,b}^{HIS}=\frac{\gamma^{0.2}}{\gamma^{0.3}+0.5(\alpha^{0.3}+\beta^{0.3})},
\end{align}
Here, the exponent parameters $\{0.2$, $0.3\}$ and the scalar parameter $0.5$ are designed empirically, inspired by the unigram distribution with 3/4rd power in word2vec \cite{mikolov2013distributed}. 

The initial scalars $\alpha,\beta,\gamma$ of the HIS Similarity will be used as labels in the training of sense spectra, which is introduced in Subsection 3.3.

\subsection{Three basic synset similarities}
There are three basic measurements on the similarity between two noun or verb synsets in WordNet: The Shortest Path Similarity, Leacock-Chodorow Similarity and Wu-Palmer Similarity \cite{WordNet_Similarities}. They are ``basic" since they only require the hypernym-hyponym relationship between two synsets \cite{Information_content}, which is the same as the HIS Similarity. Hence, they are used as the comparisons to our model. 

$\bullet$ $\mathbf{Shortest \ Path \ Similarity}$: All the noun synsets share the same root hypernym $entity.n.01$. But there may be no common hypernym between two verb synsets. So, a fake root synset $root.v.01$ is added to the verb synsets. 

Then, for any two noun or verb synsets $a$ and $b$, there is always a hypernym-hyponym path connecting them through a common hypernym of them. And there is a shortest path among all these paths, whose length is denoted as $l_{a,b}$. The Shortest Path Similarity between synsets $a$ and $b$ is then defined to be $\mathcal{K}_{a,b}^{s.p.}=1/l_{a,b}$, which is between 0 and 1.

$\bullet$ $\mathbf{Leacock\!-\! Chodorow \  Similarity}$: The $depth$ of a noun or verb synset $a$, denoted as $d_{a}$,  is defined to be the length of the shortest path from $a$ to the root synset ($entity.n.01$ for noun and $root.v.01$ for verb). That is, $d_a=l_{a,root}$.

Then, for two synsets $a$ and $b$, the Leacock-Chodorow (LCH) Similarity is defined as
$$\mathcal{K}_{a,b}^{LCH}=-\log\frac{l_{a,b}}{2\cdot \max\{d_a,d_b\}}.$$

$\bullet$ $\mathbf{Wu\!-\!Palmer \  Similarity}$: For two noun or verb synsets $a$ and $b$, their $Least \ Common$ $Subsumer$ (LCS) is the common hypernym of $a$ and $b$ with the largest depth. We use $\hat{h}_{a,b}$ or simply $\hat{h}$ to denote the LCS of synsets $a$ and $b$. That is, $\hat{h}=\max_{h\in S_{a\cap b}}\{d_h\}$. Then, the Wu-Palmer (WP) Similarity between synsets $a$ and $b$ is defined as 
$\mathcal{K}_{a,b}^{WP}=2d_{\hat{h}}/(d_a+d_b)$.



\subsection{Sense spectrum}

Suppose $v_a$ and $v_b$ are the embedding vectors of synsets $a$ and $b$ respectively. Then, we use the ``overlapping" between $v_a$ and $v_b$ to represent the ``commonness" between synsets $a$ and $b$. The overlapping of two vectors is measured dimension-wise: Suppose ${v_a}_i$ and ${v_b}_i$ are the elements in the $i$'th dimension of $v_a$ and $v_b$ respectively. We use ${v_a}_i\cap {v_b}_i$ to represent the overlapping between ${v_a}_i$ and ${v_b}_i$. Then, if ${v_a}_i$ and ${v_b}_i$ have the same sign (i.e., both of them are positive or both are negative), ${v_a}_i\cap {v_b}_i$ will equal to the one of ${v_a}_i$ and ${v_b}_i$ with the smaller absolute value. If ${v_a}_i$ and ${v_b}_i$ have different signs, ${v_a}_i\cap {v_b}_i$ will be zero. That is, mathematically:
\begin{align}
{v_a}_i\!\cap\! {v_b}_i\!=\!\frac{\mathrm{sgn}(\!{v_a}_i\!)\!+\!\mathrm{sgn}(\!{v_b}_i\!)}{2}\!\cdot\! \min\{|\!{v_a}_i|, |\!{v_b}_i|\}.\!\!\!
\end{align}
where $\mathrm{sgn}(x)$ is the sign function:
$$\mathrm{sgn}(x)=\begin{cases}-1, \ \ \mathrm{if} \ x<0, \\
0, \ \ \mathrm{if} \ x=0,\\
1, \ \ \mathrm{if} \ x>0.
 \end{cases}$$

Taking this operation to each dimension $i$, we can get the overlapping vector $v_a\cap v_b$ by $\{v_a\cap v_b\}_i={v_a}_i\cap {v_b}_i$. We also denote $v_a\cap v_b$ as $v_{a\cap b}$, which can be regarded as the vector representation on the ``commonness" between synsets $a$ and $b$. After obtaining $v_{a\cap b}$, the vector representation on the ``differences (uniqueness)" of synsets $a$ and $b$ is obvious: We use $v_{a/b}=v_a-v_{a\cap b}$ to represent the ``uniqueness" of synset $a$, and use $v_{b/a}=v_b-v_{a\cap b}$ to represent the ``uniqueness" of synset $b$. 

We can see that each dimension in $v_a$ and $v_b$ operates independently to form $v_{a\cap b}$, $v_{a/b}$ and $v_{b/a}$. This makes our embedding vector looks like a ``spectrum", with its dimensions to be the measurements on specific senses. In fact, this is verified by experiments, which will be discussed in Section 5.  
As a result, we call our synset embedding the $\mathbf{Sense \  Spectrum}$. For any two synsets $a$ and $b$, we call $v_{a\cap b}$ the Commonness Spectrum, and we call $v_{a/b}$, $v_{b/a}$ the Uniqueness Spectra. Figure 2 provides a clear exhibition on how to obtain $v_{a\cap b}$, $v_{a/b}$ and $v_{b/a}$ based on the initial spectra $v_a$ and $v_b$.
\begin{figure}[H]
\centering
\includegraphics[width=8cm]{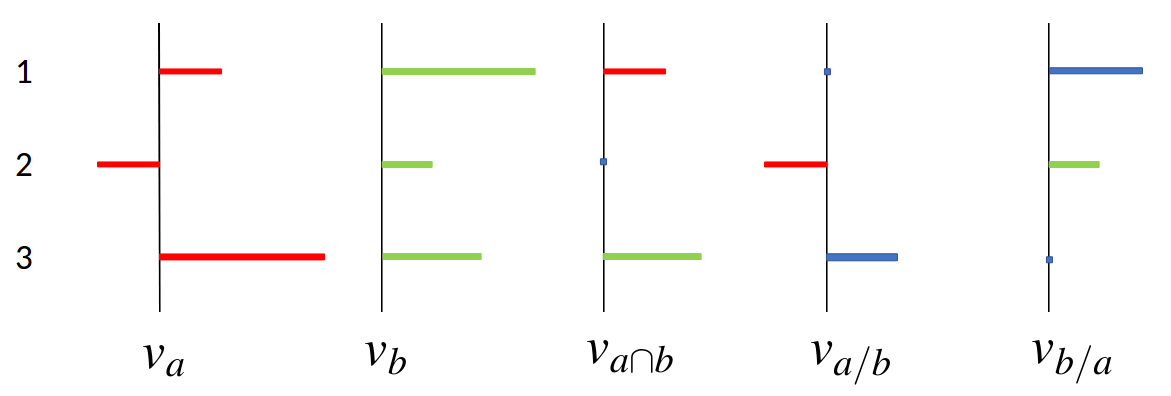}
\caption{(To be viewed in color) The commonness spectrum $v_{a\cap b}$ as well as the uniqueness spectra $v_{a/b}$ and $v_{b/a}$, based on the initial sense spectra $v_a$ and $v_b$.}
\end{figure}

In Figure 2, we suppose a spectrum vector is dimension-three. We show $v_a$ in red and $v_b$ in green. According to our overlapping method, the dimension 1 of $v_{a\cap b}$ is the same as that of $v_a$, which is also in red. Similarly, dimension 3 of $v_{a\cap b}$ is the same as that of $v_b$, which is in green. But $v_a$ and $v_b$ are not overlapped in dimension 2, making that of $v_{a\cap b}$ to be zero. Hence, dimension 2 of $v_{a/b}$ and $v_{b/a}$ shall remain the same as in $v_a$ and $v_b$ respectively, since no cancellation is made from $v_{a\cap b}$. Finally, dimension 1 of $v_{a/b}$ and dimension 3 of $v_{b/a}$ are cancelled to zero. But dimension 3 of $v_{a/b}$ and dimension 1 of $v_{b/a}$ are partially cancelled, which is in blue.

It is then only straightforward to figure out that the three spectra $v_{a\cap b}$, $v_{a/b}$ and $v_{b/a}$ coincide with the initial HIS scalars $\gamma$, $\alpha$ and $\beta$: The commonness spectrum $v_{a\cap b}$ coincides with the scalar $\gamma$, while the two uniqueness spectra $v_{a/b}$ and $v_{b/a}$ coincide with scalars $\alpha$ and $\beta$, respectively. Hence, our training algorithm is as simple as:
\begin{align}
|\!|v_{a\cap b}|\!|_1=\gamma, \ \ |\!|v_{a/b}|\!|_1=\alpha, \ \ |\!|v_{b/a}|\!|_1=\beta,
\end{align}
where $|\!|\cdot|\!|_1$ is the $L_1$ norm of a vector \cite{vector_norm}.

We will show in the next section that after training, the hypernym-hyponym relationship between two noun or verb synsets $a$ and $b$ is preserved in their corresponding spectra $v_a$ and $v_b$. 

\section{Evaluation}
In this section, we show by experimental results that the Hypernym Intersection Similarity outperforms the three basic similarity measurements in WordNet. And we will show that Sense Spectra indeed capture the structures of the hypernym-hyponym relationship in WordNet.

\subsection{The performance of HIS}
To estimate the performance of HIS Similarity, we use the dataset SimLex-999 \cite{SimLex-999}, which contains 666 noun pairs, 222 verb pairs and 111 adjective pairs. Each pair of words in SimLex-999 is scored from 0 to 10: The higher the score is, the more similar the two words in that pair should be. All the scores are given manually by native English speakers. Table 1 provides a brief exhibition on the noun and verb pairs in SimLex-999.

\begin{table}[htbp]
\caption{The noun and verb pairs as well as their corresponding scores in SimLex-999}
\begin{center}
\begin{tabular}{|c|c|c||c|c|c|}
\hline
\multicolumn{2}{|c|}{Noun pairs}&$\!$Score$\!$&\multicolumn{2}{|c|}{Verb pairs}&$\!$Score$\!$\\
\hline
book&text&6.35&listen&$\!$hear$\!$&8.17\\
\hline
night&day&1.88&go&$\!$come$\!$&2.42\\
\hline
$\!$belief$\!$&$\!$flower$\!$&0.40&$\!$spend$\!$&save&0.55\\
\hline
$\cdots$&$\cdots$&$\cdots$&$\cdots$&$\cdots$&$\cdots$\\
\hline
\end{tabular}
\label{tab1}
\end{center}
\end{table} 

However, there may be more than one synset related to a word \cite{navigli2009word}. For example, there are 11 noun synsets related to the word ``book", including $book.n.01$ (a written work or composition that has been published), $book.n.02$ (physical objects consisting of a number of pages bound together), $bible.n.01$ (the sacred writings of the Christian religions), etc. So, we need to first choose the correct synsets for each word pair in SimLex-999.  

For a word pair $(w_1, w_2)$, suppose there are $M$ synsets $\{s_1,\cdots,s_M\}$ related to $w_1$, and $N$ synsets $\{s'_1,\cdots,s'_N\}$ related to $w_2$. Then, there are $M\times N$ possible combinations of synsets for the word pair $(w_1, w_2)$: $\{(s_1,s'_1),\cdots,(s_1,s'_N),\cdots,(s_M,s'_1),\cdots,(s_M,s'_N)\}$. We compute the HIS Similarity $\mathcal{K}_{s_m,s'_n}^{HIS}$ by formula (1) on each synset combination $(s_m,s'_n)$, and then choose the combination $(\hat{s}_m,\hat{s}'_n)$ with the maximal similarity score $\mathcal{K}_{\hat{s}_m,\hat{s}'_n}^{HIS}$. After that, $(\hat{s}_m,\hat{s}'_n)$ is regarded as the correct synset choice for the word pair $(w_1, w_2)$, and $\mathcal{K}_{\hat{s}_m,\hat{s}'_n}^{HIS}$ is regarded as the similarity score of $(w_1, w_2)$ under the HIS Similarity, denoted as $\mathcal{K}_{w_1,w_2}^{HIS}$. 

Finally, suppose $\{(w_1^k,w_2^k)\}_{k=1}^K$ is a specific set of word pairs in SimLex-999 (say, all the noun pairs). For each word pair $(w_1^k,w_2^k)$, suppose $\mathcal{S}_k$ is the manually given similarity score in SimLex-999, and $\mathcal{K}_k:=\mathcal{K}_{w_1^k,w_2^k}^{HIS}$ is the similarity score under HIS Similarity. Then, we compute the Spearman's correlation \cite{Handbook_bio_stat} between $\{\mathcal{S}_k\}_{k=1}^K$ and $\{\mathcal{K}_k\}_{k=1}^K$ as:
\begin{align}
\rho=\frac{\sum_{k=1}^K(\mathcal{S}_k-\overline{\mathcal{S}})(\mathcal{K}_k-\overline{\mathcal{K}})}{\sqrt{\sum_{k=1}^K(\mathcal{S}_k-\overline{\mathcal{S}})^2\sum_{k=1}^K(\mathcal{K}_k-\overline{\mathcal{K}})^2}},
\end{align}
where $\overline{\mathcal{S}}$ and $\overline{\mathcal{K}}$ are the averages of $\{\mathcal{S}_k\}_{k=1}^K$ and $\{\mathcal{K}_k\}_{k=1}^K$ respectively. This Spearman's correlation $\rho$ is then the estimation on the performance of the HIS Similarity. A higher Spearman's correlation here means that the language model can handle the semantic meanings of words more like humans do \cite{interpret_Spearmans_corr}. 

In order to obtain comparisons, we apply the same process onto the Shortest Path Similarity, Leacock-Chodorow (LCH) Similarity and Wu-Palmer (WP) Similarity. That is, we replace the similarity score $\mathcal{K}_{s_m,s'_n}^{HIS}$ with $\mathcal{K}_{s_m,s'_n}^{s.p.}$, $\mathcal{K}_{s_m,s'_n}^{LCH}$ and $\mathcal{K}_{s_m,s'_n}^{WP}$ as described in Section 3.2 to get the corresponding Spearman's correlation $\rho_{s.p.}$, $\rho_{LCH}$ and $\rho_{WU}$, respectively. Moreover, we work on three different sets of word pairs in SimLex-999: only noun pairs, only verb pairs, or combining both noun and verb ones. Results are shown in Table 2.

\begin{table}[htbp]
\caption{The Spearman's correlations obtained by performing each similarity measurement on different sets of word pairs in SimLex-999.}
\begin{center}
\begin{tabular}{|c|c|c|c|}
\hline
\backslashbox{$\!\!\!\!$Model$\!\!\!\!\!\!\!\!$}{$\!\!\!\!\!\!\!$Group$\!\!\!$}&$\!$Noun pairs$\!$&$\!$Verb pairs$\!$&Both\\
\hline
HIS&$\mathbf{61.12}$&$\mathbf{48.38}$&$\mathbf{55.98}$\\
\hline
$\!\!$Shortest Path$\!\!$&58.38&39.20&51.96\\
\hline
LCH&58.38&39.20&54.92\\
\hline
WP&55.00&37.84&48.82\\
\hline
\end{tabular}
\label{tab2}
\end{center}
\end{table} 

We can see that the HIS Similarity achieves the highest Spearman's correlation on all the three sets of word pairs. To be specific, on the verb pairs, the HIS Similarity outperforms the other three similarity measurements by 10 percent, which is a significant improvement. 

 Therefore, we claim that the HIS Similarity captures the hypernym-hyponym relationship in WordNet better than the three basic similarities do. 
Hence, it is meaningful to use the initial HIS scalars $\alpha$, $\beta$, $\gamma$ as labels to train our sense spectra, whose performance is given in the following subsection.

\subsection{The performance of sense spectra}
Again, we note that it is meaningless to compare a noun synset with a verb one. So, the noun and verb spectra are generated and trained independently: There are 82,115 noun synsets, whose spectra are generated as $v_1,\cdots,v_{82115}$; And there are 13,767 verb synsets, whose spectra are $v'_1,\cdots,v'_{13767}$. We always set the dimension of a spectrum to be $D=200$ for both noun and verb synsets.

When training the sense spectra, we use TensorFlow in Python \cite{tensorflow}. We shall first introduce our methods of implementations. Then, we shall introduce our specific strategy on how to build a training batch. Finally, meaningful testing results will be given.

\subsubsection{Implementation issues}
When computing the dimension-wise overlapping ${v_a}_i\cap {v_b}_i$, we realize that it is difficult to perform formula (2) directly in TensorFlow. This is because errors cannot path through the sign function $\mathrm{sgn}(x)$ by back propagation \cite{Back_propagation}. Besides, there is no necessary to generate $v_{a\cap b}$ by each dimension in practice. So, we use a formula evolved with the rectifier function \cite{ReLU} $\mathrm{ReLU}(x)=\max\{0,x\}$ to compute $v_{a\cap b}$ directly: 
\begin{align*}
v_{a\cap b}=\min&(\mathrm{ReLU}(v_a),\mathrm{ReLU}(v_b))\\
&-\min(\mathrm{ReLU}(-v_a),\mathrm{ReLU}(-v_b))
\end{align*}

To be specific, suppose the dimension of a spectrum vector is $D$. We first apply $t\!f.concat$ to concatenate the rectified vectors $\mathrm{ReLU}(v_a)$ and $\mathrm{ReLU}(v_b)$ along each dimension, which returns a $D\times 2$ tensor (matrix) $c_+$. After that, we apply $t\!f.reduce\_min$ to get the minimum value on each dimension of $c_+$, which returns a $D$ dimensional vector $t_+$. Similarly, we can get $c_-$ and $t_-$ with respect to $\mathrm{ReLU}(-v_a)$ and $\mathrm{ReLU}(-v_b)$. Then, we have that $v_{a\cap b}=t_+-t_-$.

The formulas to obtain $v_{a/b}$ and $v_{b/a}$ in practice are much more straightforward:
\begin{align*}
v_{a/b}=&\mathrm{ReLU}(\mathrm{ReLU}(v_a)-\mathrm{ReLU}(v_b))\\
&-\mathrm{ReLU}(\mathrm{ReLU}(-v_a)-\mathrm{ReLU}(-v_b)),\\
v_{b/a}=&\mathrm{ReLU}(\mathrm{ReLU}(v_b)-\mathrm{ReLU}(v_a))\\
&-\mathrm{ReLU}(\mathrm{ReLU}(-v_b)-\mathrm{ReLU}(-v_a)).
\end{align*}

After that, we compute the $L_1$ norm of a $D$ dimensional vector $v$ as $|\!|v|\!|_1=\sum_{d=1}^D|v_d|=t\!f.reduce\_sum(t\!f.abs(v)).$

Finally, suppose $|\!|v_{a/b}|\!|_1=\hat{\alpha}$, $|\!|v_{b/a}|\!|_1=\hat{\beta}$ and $|\!|v_{a\cap b}|\!|_1=\hat{\gamma}$. Applying the initial HIS scalars $\alpha$, $\beta$, $\gamma$ as labels, we complete the training by minimizing the error $|\alpha-\hat{\alpha}|+|\beta-\hat{\beta}|+|\gamma-\hat{\gamma}|$ via AdamOptimizer \cite{AdamOptimization}.


\subsubsection{Batch formation strategies}
Each batch in our model consists of a synset pair $(a,b)$, which is formed in three different ways:

$\circ$ The direct hypernym pair: After choosing a synset $a$ randomly, we pick its direct hypernym $h_a$ to form a pair $(a,h_a)$. If there are more than one direct hypernyms for the synset $a$, we shall choose one of them randomly.

$\circ$ The semantic sense related pair: As we mentioned in the introduction, each synset is related to a specific semantic sense. There are 67,176 noun semantic senses and 7,440 verb semantic senses that have more than one related synsets. In the training, we shall randomly pick one semantic sense and randomly choose two of its related synsets to form a semantic sense related pair. That is, suppose we get the semantic sense $S$ and its related synsets $\{s_1,\cdots, s_k\}$. Then, we randomly pick two synsets $\hat{s}_1,\hat{s}_2$ from $\{s_1,\cdots, s_k\}$ to from the pair.

$\circ$ Random pair: We choose two synsets $a$ and $b$ randomly to form the pair.

 We have $T$ pairs built in each of these three ways. So, our total batch size is $3T$. We always set $T=100$ in our training. And again, we note that noun and verb synset pairs are formed independently. 

\subsubsection{Testing results} 
After training, we look up the three closest spectra for each spectrum under the HIS Similarity. That is, for a spectrum vector $v_a$ related to the synset $a$, we compute $|\!|v_{a/b}|\!|_1=\hat{\alpha}$, $|\!|v_{b/a}|\!|_1=\hat{\beta}$ and $|\!|v_{a\cap b}|\!|_1=\hat{\gamma}$ with respect to every else noun (or verb) spectrum $v_b$. Then, we apply formula (1) on each scalar set $(\hat{\alpha}, \hat{\beta}, \hat{\gamma})_b$ to find the three synsets $b_1$, $b_2$ and $b_3$ that provide the maximal values $\mathcal{K}_{a,b}^{HIS}$. Again, we note that this procedure is performed on the noun and the verb spectra independently. Some results are shown in Table 3.

\begin{table*}[htbp]
\caption{The three closest spectra for each sense spectrum under the HIS Similarity.}
\begin{center}
\begin{tabular}{|c||c|c||c|c||c|c|}
\hline
Synset $a$&Synset $b_1$&$\!\!\mathcal{K}_{a,b_1}^{HIS}\!\!$&Synset $b_2$&$\!\!\mathcal{K}_{a,b_2}^{HIS}\!\!$&Synset $b_3$&$\!\!\mathcal{K}_{a,b_3}^{HIS}\!\!$\\
\hline
$trade.n.01$&$\!fair\_trade.n.02\!$&8.04&$\!fair\_trade.n.01\!$&8.03&$free\_trade.n.01$&7.59\\
\hline
$\!finance.n.01\!$&$flotation.n.02$&8.98&$banking.n.02$&8.90&$\!\!high\_finance.n.01\!\!\!$&8.88\\
\hline
$war.n.01$&$jihad.n.01$&7.06&$hot\_war.n.01$&7.02&$world\_war.n.01$&7.01\\
\hline
$vent.n.01$&$\!smoke\_hole.n.01\!$&6.80&$bunghole.n.02$&6.32&$air\_hole.n.02$&6.29\\
\hline
$receipt.n.02$&$\!\!bill\_of\_lading.n.01\!\!\!$&4.99&$\!pawn\_ticket.n.01\!$&4.87&$stub.n.03$&4.55\\
\hline
$\!\!myosotis.n.01\!\!$&$\!\!cynoglossum.n.01\!\!$&6.22&$\!\!genus\_martynia.n.01\!\!$&6.21&$\!\!\!physostigma.n.01\!\!\!$&6.20\\
\hline
$\!\!cough.v.01\!\!$&$\!\!hack.v.08\!\!$&1.91&$\!\!\!clear\_the\_throat.v.01\!\!\!$&1.87&$\!\!\!expectorate.v.02\!\!\!$&0.93\\
\hline
$\!\!laugh.v.01\!\!$&$\!\!snicker.v.01\!\!$&0.98&$\!\!\!break\_up.v.19\!\!\!$&0.97&$\!\!\!cackle.v.03\!\!\!$&0.97\\
\hline
$\!\!coach.v.01\!\!$&$\!\!condition.v.01\!\!$&4.00&$\!\!\!mentor.v.01\!\!\!$&3.93&$\!\!\!reinforce.v.02\!\!\!$&3.92\\
\hline
$\!\!propose.v.01\!\!$&$\!\!submit.v.02\!\!$&2.99&$\!\!\!recommend.v.01\!\!\!$&2.93&$\!\!\!advance.v.02\!\!\!$&2.93\\
\hline
\end{tabular}
\label{tab3}
\end{center}
\end{table*}

 Table 3 contains the synsets related to both commonly used words and specific terminologies. To be specific, synsets $cynoglossum.n.01$, $genus\_martynia.n.01$ and $physostigma.n.01$ represents three different genera of plants, to which $myosotis.n.01$ belongs; The synset $stub.n.03$ is defined as ``a torn part of a ticket returned to the holder as a receipt"; And the synset $break\_up.v.19$ means ``laugh unrestrainedly". By these examples, we can see that sense spectra with similar meanings (corresponding to their synsets) are clustered together under the HIS Similarity, which is similar to the performance of word2vec \cite{mikolov2013distributed}. This result shows that sense spectra to some extend preserve the hypernym-hyponym relationship in WordNet after training. 
 
However, one may ask: To what extend, or how precise, can sense spectra preserve the hypernym-hyponym relationship? 
In order to answer this question, we pick a pair of synsets and their corresponding spectra $(a,b,v_a,v_b)$, and then compare the initial HIS scalars $(\alpha, \beta,\gamma)_{(a,b)}$ with the spectrum-based HIS scalars $(\hat{\alpha}\!\!=\!\!|\!|v_{a/b}|\!|_1,\ \ \hat{\beta}\!\!=\!\!|\!|v_{b/a}|\!|_1,\ \ \hat{\gamma}\!=\!\!|\!|v_{a\cap b}|\!|_1\!)$ by
\begin{align}
R_{(a,b)}=\frac{|\alpha-\hat{\alpha}|+|\beta-\hat{\beta}|+|\gamma-\hat{\gamma}|}{\alpha+\beta+\gamma}.
\end{align}
The numerator of formula (5) represents the error made by the sense spectra, while the denominator represents the magnitude of the initial HIS scalars. Hence, the smaller $R_{(a,b)}$ is, the less important the error is comparing to the initial HIS scalars, and hence the more precise sense spectra $v_a,v_b$ can capture the hypernym-hyponym relationship between their corresponding synsets $a,b$.

We perform formula (1) on SimLex-999 dataset. That is, we obtain the synset pair $(\hat{s}_m,\hat{s}'_n)$ for each noun and verb pair $(w_1,w_2)$ in SimLex-999, and then compute $R_{(\hat{s}_m,\hat{s}'_n)}$ on each $(\hat{s}_m,\hat{s}'_n, v_{\hat{s}_m}, v_{\hat{s}'_n})$. The distribution of the pairs $(\hat{s}_m,\hat{s}'_n, v_{\hat{s}_m}, v_{\hat{s}'_n})$ based on the values $R_{(\hat{s}_m,\hat{s}'_n)}$ is observed by the histograms in Figure 3 and the statistical results in Table 4.

\begin{figure}[H]
\centering
\includegraphics[width=8cm]{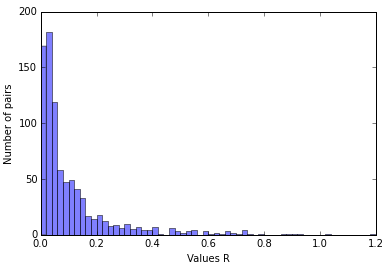}
\caption{The histogram distribution of the pairs $(\hat{s}_m,\hat{s}'_n, v_{\hat{s}_m}, v_{\hat{s}'_n})$ based on the Values $R_{(\hat{s}_m,\hat{s}'_n)}$.}
\end{figure} 

\begin{table}[htbp]
\caption{The statistical distribution of the pairs $(\hat{s}_m,\hat{s}'_n, v_{\hat{s}_m}, v_{\hat{s}'_n})$ based on the Values $R_{(\hat{s}_m,\hat{s}'_n)}$.}
\begin{center}
\begin{tabular}{|c|c|c|c|c|}
\hline
$\!\!\!$Range of $R_{(\hat{s}_m,\hat{s}'_n)}\!\!\!$&$\!\!\!<0.05\!\!\!$&$\!\!\!<0.1\!\!\!$&$\!\!\!<0.2\!\!\!$&$\!\!\!<0.3\!\!\!$\\
\hline
$\!\!\!$Percentage of Pairs $\!\!\!\!\!$&$\!\!\!47.81\!\!\!$&$\!\!\!66.24\!\!\!$&$\!\!\!83.99\!\!\!$&$\!\!\!90.09\!\!\!$\\
\hline
\end{tabular}
\label{tab4}
\end{center}
\end{table} 

We can see from Figure 3 and Table 4 that more than $90\%$ of the pairs $(\hat{s}_m,\hat{s}'_n, v_{\hat{s}_m}, v_{\hat{s}'_n})$ have a value $R_{(\hat{s}_m,\hat{s}'_n)}$ significantly less than one. That is, in most SimLex-999 pairs $(\hat{s}_m,\hat{s}'_n, v_{\hat{s}_m}, v_{\hat{s}'_n})$, the sense spectra $v_{\hat{s}_m}$, $v_{\hat{s}'_n}$ capture the hypernym-hyponym relationship between the synsets $\hat{s}_m$, $\hat{s}'_n$ precisely. Taking the authority and complexity of the SimLex-999 dataset, we claim that in general, our sense spectra capture the hypernym-hyponym relationships among their corresponding synsets precisely. To the best of our knowledge, this is the first time that low dimensional embeddings can preserve the semantic relationships among synsets in WordNet.

Finally, in Figure 4, we plot the spectra for the synsets in Table 3 that are related to $trade.n.01$ (row 1) and $cough.v.01$ (row 7). Different from vertical spectra in Figure 2, we plot horizontal spectra here to save spaces.

\begin{figure*}[htbp]
\caption{Spectra for the synsets related to $trade.n.01$ and $cough.v.01$ in Table 3.}      
\begin{minipage}[t]{0.5\linewidth}
\centering
\includegraphics[width=3 in]{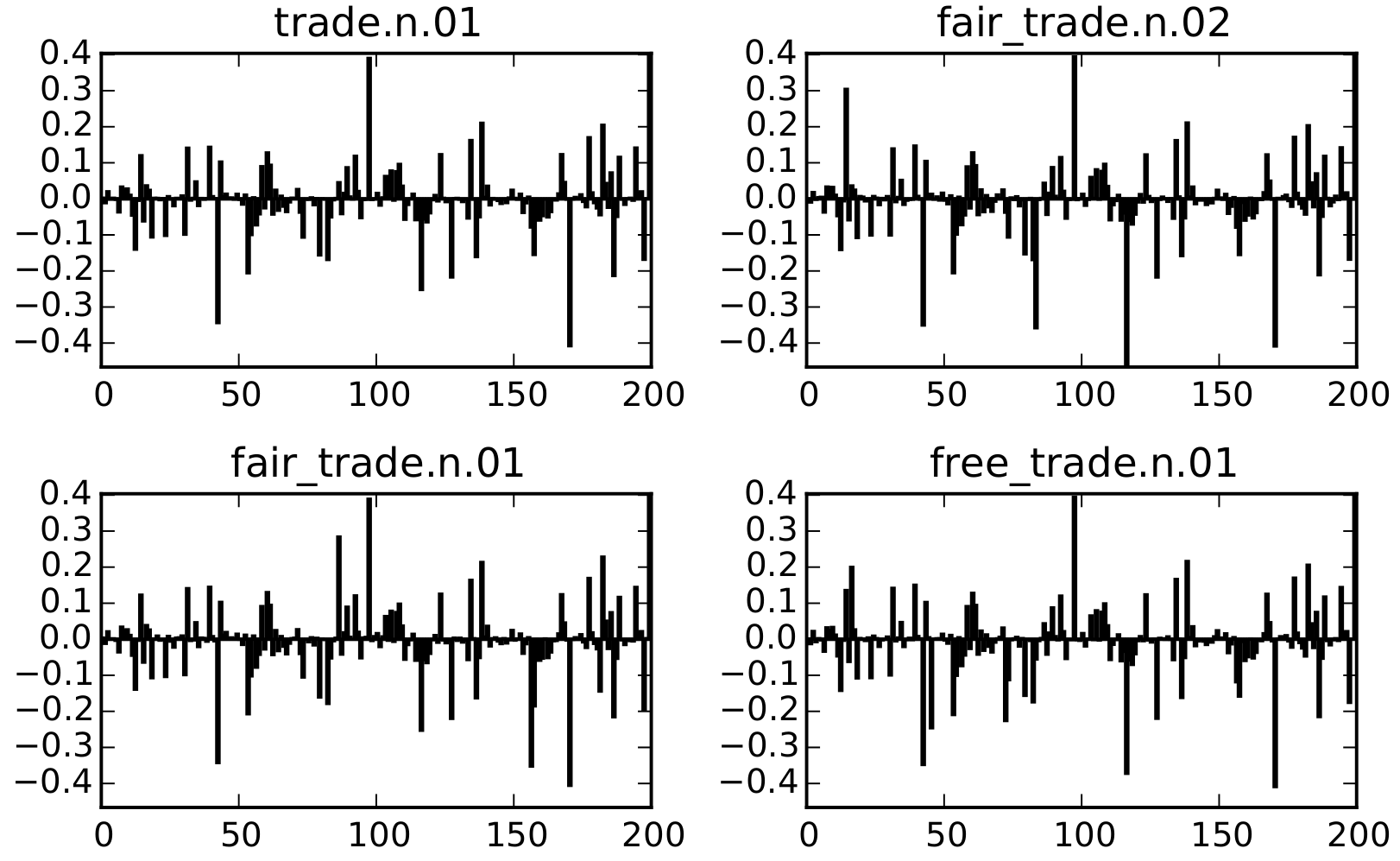}   
\label{fig:side:a}
\end{minipage}%
\begin{minipage}[t]{0.5\linewidth}
\centering
\includegraphics[width=3 in]{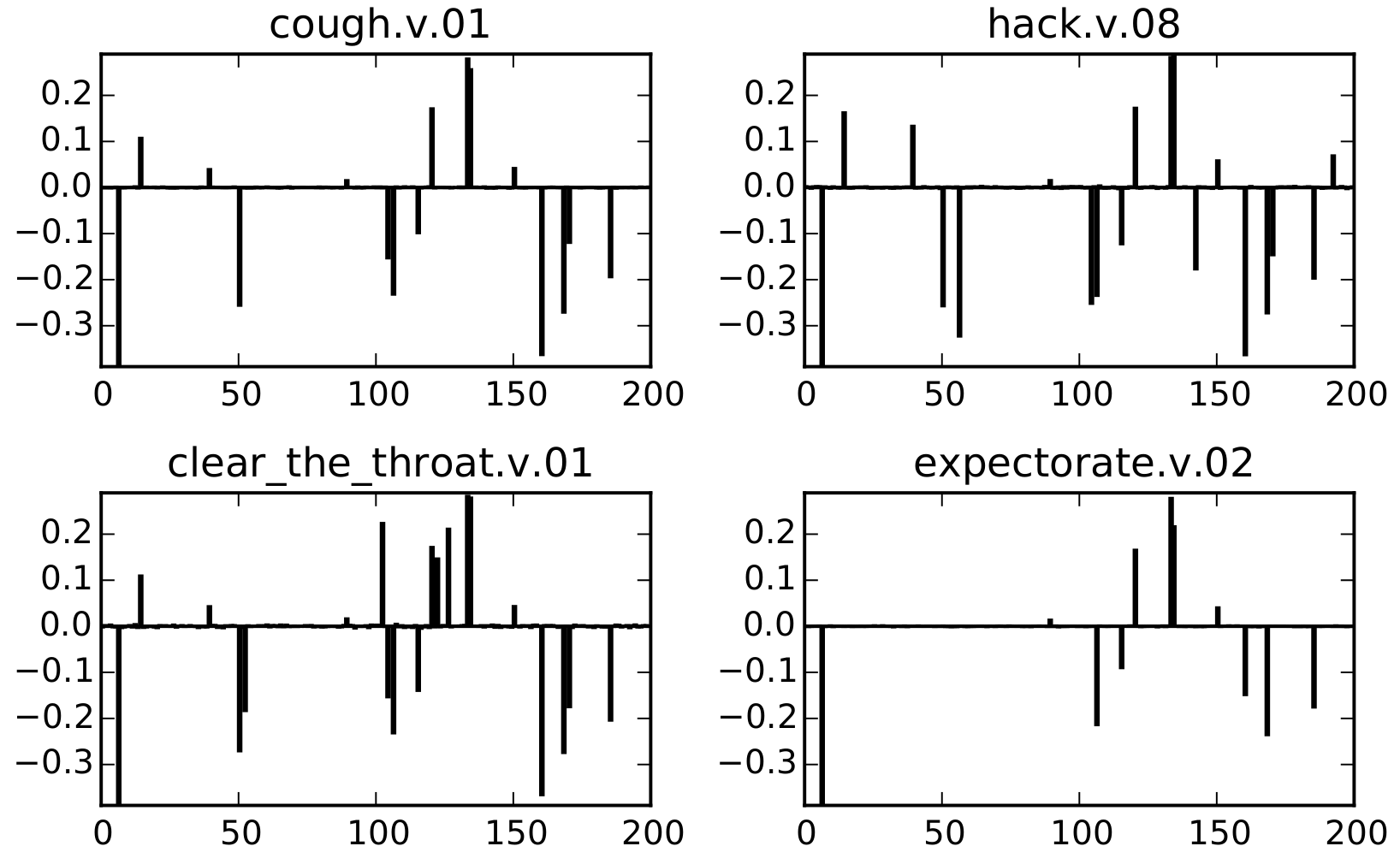}     
\label{fig:side:b}
\end{minipage}
\end{figure*} 

We can see from Figure 4 that verb spectra are in general sparser than noun spectra. This is because the hypernym-hyponym relationships among verb synsets are more concise comparing to those among noun synsets. Hence, fewer dimensions in a spectrum are enough to preserve the semantic information. 
In addition, spectra with similar meanings (corresponding to their synsets) also have similar distributions across dimensions. That is, in most dimensions, spectra with similar meanings tend to have the same sign and the same magnitude. These phenomena are in fact meaningful, which will be discussed in the next section.




$\mathbf{Reproducability}$: Our code can be accessed via github. The link will be provided upon acceptance.

\section{Discussions}
In this section, we shall further discuss the meaning behind our experimental results, based on which we shall describe about the potential applications of the sense spectra.

\subsection{Building hierarchical language model based on sense spectra}
As we mentioned in the previous section, sense spectra with similar meanings tend to have the same sign and magnitude in most dimensions. However, since the number of dimensions in a spectrum is far less than the number of (noun or verb) synsets, it is impossible for each dimension to preserve semantic information independently. 
As a result, we can conclude that specific combinations of dimensions in a sense spectrum work together to preserve specific semantic information. That is, there exists structures related to semantic senses among the dimensions in sense spectra, indicating the name Sense Spectrum is fair and genuine. 


Then, combining with text training corpus \cite{Text_corpus}, it is possible to build hierarchical language model based on the structures in sense spectra. For example, we may first group together the dimensions that co-occur frequently across noun or verb spectra. These dimension groups should be highly related to the dimension combinations preserving specific semantic information.
Then, for each word in the training corpus, we may find all the related synsets and put the corresponding spectra into a list. After that, we will have a list of the possible spectra for each word in the training corpus. Finally, based on this, we may discover the ``groups of dimension groups." That is, we further group together the dimension groups that co-occur frequently in the training corpus. In this way, a hierarchical language model with explicit upper layer units can be built. 

\subsection{Combining sense spectra with word embeddings}
Now that sense spectra are low dimensional and dense, we can directly concatenate them to the pre-trained word embeddings. Similar to the above discussion, for each word $w$ in the training corpus, we may find its related synsets. Then, we pick out the corresponding spectra of these synsets and perform the average summation over the spectra to get a summation vector. Finally, we concatenate the summation vector to the pre-trained embedding vector of the word $w$.  

In this way, the embedding vectors now contain not only the contextual information obtained from the corpus-based training, but also the semantic relationship information obtained from the knowledge-based training \cite{Boom_2016_rep}. We believe that such word embeddings are promising for tasks such as Word Sense Disambiguation (WSD) \cite{Yarowsky2000} and Outlier Detection \cite{Outlier_Analysis_Aggarwal}, where information about semantic relationships are highly demanded. This is being investigated.

\section{Conclusion}
We provide sense spectra, which are the first dense and low-dimension embeddings for the noun and verb synsets in WordNet that preserve the hypernym-hyponym relationships among the synsets.  We train sense spectra by HIS Similarity, which is a similarity measurement describing the ``commonness" and ``uniqueness" between two noun or verb synsets in WordNet. 

Results show that the HIS Similarity outperforms the three basic similarity measurements in WordNet on the SimLex-999 noun and verb pairs, and sense spectra do preserve the hypernym-hyponym relationship among synsets precisely. Novel applications built on sense spectra are described and are being actively explored.


\bibliography{custom}
\bibliographystyle{IEEEtran}

\end{document}